\pdfoutput=1 
\documentclass{article}

\usepackage[final,nonatbib]{neurips_2019}

\usepackage[utf8]{inputenc} 
\usepackage[T1]{fontenc}    
\usepackage[hidelinks]{hyperref}       
\usepackage{url}            
\usepackage{booktabs}       
\usepackage{amsfonts}       
\usepackage{nicefrac}       
\usepackage{microtype}      
\usepackage{xspace}
\usepackage[dvipdfmx]{graphicx}
\usepackage{subcaption}
\usepackage[normalem]{ulem}

\usepackage{algorithm}
\usepackage[]{algpseudocode}
\usepackage{xcolor}
\usepackage{kotex}
\usepackage[most]{tcolorbox}

\usepackage{bmpsize}

\newcommand{\hp}[0]{hyper-parameter\xspace}
\newcommand{\hps}[0]{hyper-parameters\xspace}
\newcommand{\Hp}[0]{Hyper-parameter\xspace}

\newcommand{\hpo}[0]{hyper-parameter optimization\xspace}

\newcommand{\hpoa}[0]{hyper-parameter optimization algorithm\xspace}
\newcommand{\Hpoa}[0]{Hyper-parameter optimization algorithm\xspace}
\newcommand{\hpoas}[0]{hyper-parameter optimization algorithms\xspace}
\newcommand{\study}[0]{study\xspace}
\newcommand{\studies}[0]{studies\xspace}
\newcommand{\trial}[0]{trial\xspace}
\newcommand{\trials}[0]{trials\xspace}
\newcommand{\stage}[0]{stage\xspace}
\newcommand{\stages}[0]{stages\xspace}

\newcommand{\searchspace}{search space\xspace}
\newcommand{\eat}[1]{}

\newcommand{\revise}[1]{#1}

\usepackage{listings}

\DeclareFixedFont{\ttb}{T1}{txtt}{bx}{n}{8} 
\DeclareFixedFont{\ttm}{T1}{txtt}{m}{n}{8}

\definecolor{deepblue}{rgb}{0,0,0.5}
\definecolor{deepred}{rgb}{0.6,0,0}
\definecolor{deepgreen}{rgb}{0,0.5,0}
\definecolor{gray}{rgb}{0.33,0.33,0.33}
\newcommand\pythonstyle{\lstset{
		language=Python,
		basicstyle=\ttm,
		commentstyle=\color{deepgreen},
		keywordstyle=\ttb\color{deepblue},
		emphstyle=\ttb\color{deepred},    
		frame=tb,                         
		showstringspaces=false,            %
		numbers=left,
}}

\lstnewenvironment{python}[1][]
{
	\pythonstyle
	\lstset{#1}
}
{}

\title{Stage-based Hyper-parameter Optimization \\ for Deep Learning}

\author{%
    Ahnjae Shin,
    Dong-Jin Shin,
    Sungwoo Cho,
    Do Yoon Kim,\\
    {\bf Eunji Jeong,
    Gyeong-In Yu,
    Byung-Gon Chun}\\
    Seoul National University\\
    \texttt{\{aj.shin,dongjin.shin\}@spl.snu.ac.kr},\\
    \texttt{\{sungwoocho,ddoyoon,ejjeong,gyeongin,bgchun\}@snu.ac.kr}
}

\begin{document}

\maketitle

\begin{abstract}
    As deep learning techniques advance more than ever, \hpo is the new major workload in deep learning clusters.
    Although \hpo is crucial in training deep learning models for high model performance, effectively executing such a computation-heavy workload still remains a challenge.
    We observe that numerous {\trial}s issued from existing {\hpoa}s share common \hp sequence prefixes, which implies that there are redundant computations from training the same \hp sequence multiple times.
    We propose a \stage-based execution strategy for efficient execution of {\hpoa}s.
    Our strategy removes redundancy in the training process by splitting the \hp sequences of {\trial}s into homogeneous {\stage}s, and generating a tree of {\stage}s by merging the common prefixes.
    Our preliminary experiment results show that applying \stage-based execution to {\hpoa}s outperforms the original \trial-based method, saving required GPU-hours and end-to-end training time by up to 6.60 times and 4.13 times, respectively.
\end{abstract}
\section{Introduction} \label{intro}

Deep learning (DL) models have made great leaps in various areas including image classification \cite{resnet, cifar, imagenet}, object detection \cite{yolo}, and speech recognition \cite{deepspeech, deepspeech2}.
However, such benefits come at a cost; training DL models requires heavy datasets and long computations which may take up to a week \cite{gnmt} even on a hundred of GPUs~\cite{gnmt}.
This cost becomes more significant when we take \hpo into account.
Investigating the \hp \searchspace often requires hundreds to thousands of trainings with different \hp settings \cite{massively}.
Consequently, naively running \hpo requires an exceedingly large number of GPUs, and it is crucial to explore the \hp \searchspace as efficiently as possible.
 
\begin{figure}[tb]
    \centering
    \begin{subfigure}[b]{0.57\textwidth}
       	\centering
        \includegraphics[width=1.0\textwidth]{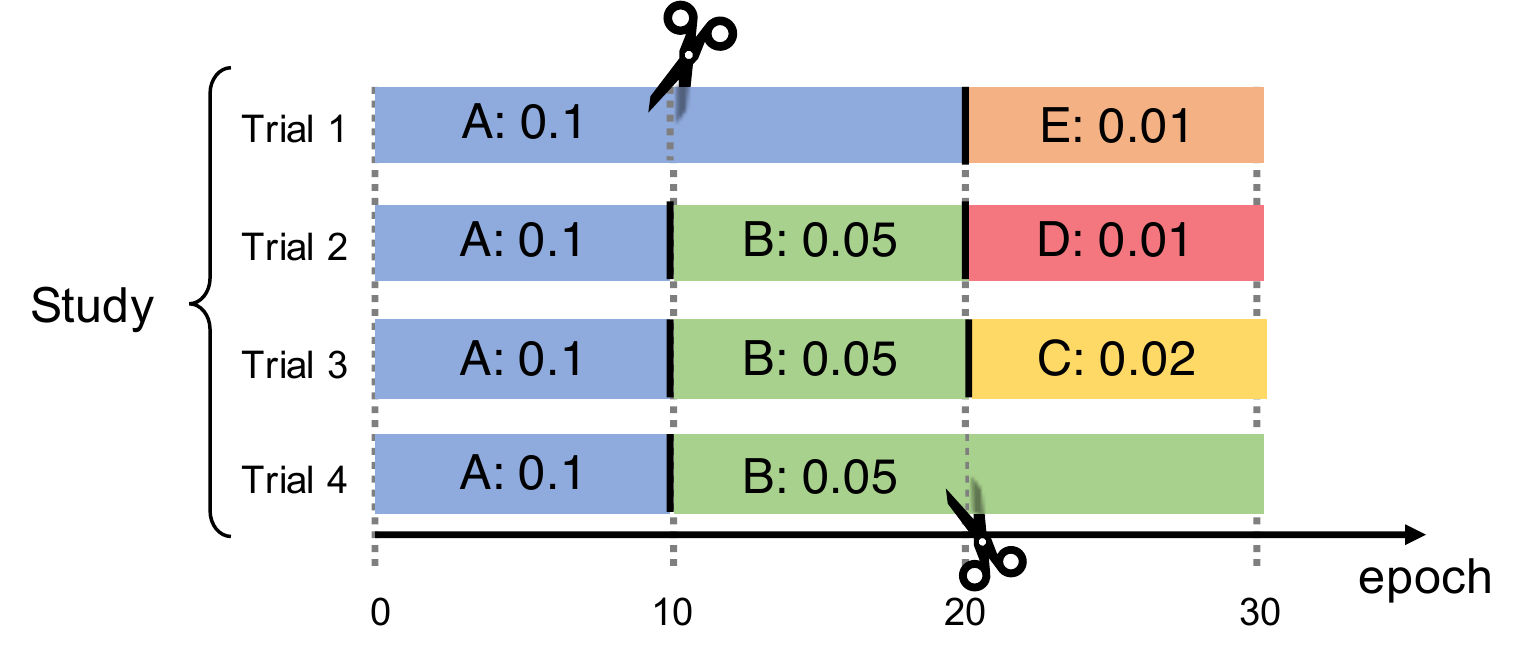}
        \centering
       	\caption{study with four trials}
       	\label{fig:hp-sequence}
    \end{subfigure}
    \hfill
    \begin{subfigure}[b]{0.42\textwidth}
        \includegraphics[width=1.0\textwidth]{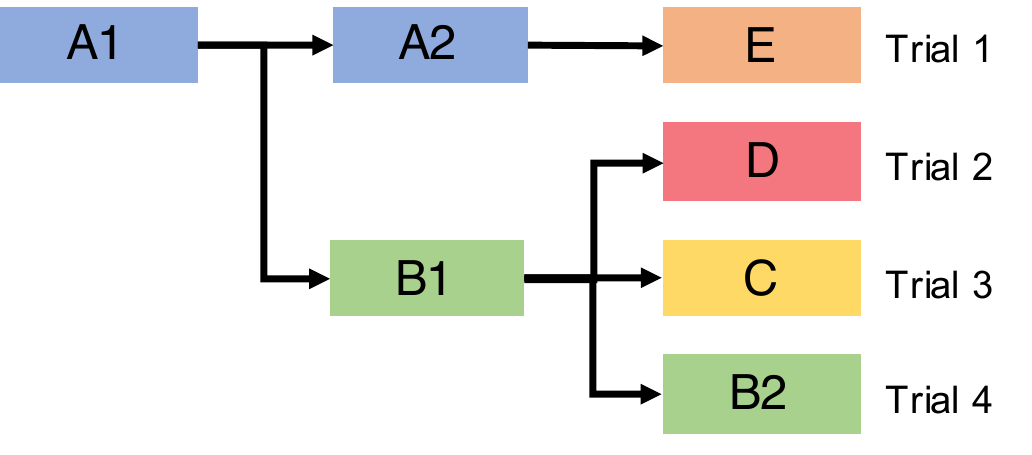}
        \vspace{0.8mm}
    	\caption{stage-tree}
    	\label{fig:stage-tree}
    \end{subfigure}
    \caption{
        Figure~\ref{fig:hp-sequence} represents a study with four different trials each with a different learning-rate sequence. Trial 1 has sequence A->E, and trial 2 has sequence A->B->D. The trials have common prefixes. Trials 2 and 3 share subsequence A in common. If we split subsequence A of trial 1 into two parts (A1 and A2), the former part is identical to the subsequence trials 2, 3, and 4 have. We can merge the common subsequences and represent the trials as a tree, as shown in Figure~\ref{fig:stage-tree}.
        Note that even if E and D have the same value, they cannot be merged and should be considered as different subsequences, because they do not share the same prefix (A1->A2 for E, A1->B1 for D).
    }
    \label{fig:main}
\end{figure}

A \hpo job trains and evaluates the target DL model multiple times, each with a different configuration. Each training sub-procedure is identified by its unique configuration. We use the term \study to refer to the job and the term \trial to refer to the training sub-procedure\footnote{The terms \study and \trial come from \textit{Vizier}\cite{vizier}.}. As an example, Figure~\ref{fig:main}(a) depicts a \study with four \trials. Each \trial has different learning-rate values. The first trial trains a DL model with 0.1 learning-rate and switches to 0.01.

Training modern DL models to reach state-of-the-art accuracy requires changing {\hp} values in the midst of training, as they target minimizing high-dimensional, non-convex loss functions.
Hence, a \hp configuration can be regarded as a sequence of values. Examples include learning-rate~\cite{resnet, batchnorm, cycliclr, super, 1hour, adam, adadelta, rmsprop, hypergradient}, drop-out ratio~\cite{elu}, optimizer~\cite{gnmt}, momentum~\cite{yellofin}, batch size~\cite{dont-decay-lr}, image augmentation parameters~\cite{pba}, training image input size~\cite{progan}, input sequence length~\cite{bert}, and network architecture parameters~\cite{progan}. 

Existing approaches for \hpo systems~\cite{raytune, vizier, mltuner, chopt} simply execute multiple \trials sequentially, or support launching multiple {\trial}s in parallel to utilize multiple GPUs and machines.
However, we observe that such \trial-based execution strategy does not exploit an important characteristic of \hpo: a \hp configuration is a \textit{sequence}, not a single \textit{value}. As in Figure~\ref{fig:hp-sequence}, a \trial is a sequence of homogeneous \textit{\stages}, where a \stage is a span of a \trial with constant {\hp} values.
For instance, the second trial is composed of three stages, and the last trial is composed of two stages. 
If the \hpo system knows where each \stage starts and ends, we can train duplicate stages only once, and reuse the computed result multiple times.
Note that, continuous-valued \hp sequences are also eligible for merging.
\revise{For instance, trials that use learning rate warmup\cite{1hour}, or cyclic learning rate strategy\cite{cycliclr} have potentially identical prefixes. When applying warmup, the learning rate linearly increases for a few steps and decreases afterward. Consequently, two different sequences can have the warmup period as a common prefix, each with a different decaying strategy.
The same logic applies to cyclic learning rates. Each cycle need not be identical to each other; usually, later cycles have smaller ranges. Hence, two different sequences may have identical cycles.}
Therefore, this observation motivates a new execution strategy for \hpo, which manages its workload based on {\stage}s, not {\trial}s. Such fine-grained execution can cut down GPU resource usage and lead to shorter end-to-end training time.

In this paper, we present a \stage-based execution strategy that seeks higher computational and resource efficiency.
With the strategy, we can exploit the characteristics of {\trial}s or {\studies} by inspecting their {\stage}s, while \trial-based execution treats {\trial}s as black boxes.
This introduces the opportunity to remove redundant computations across {\trial}s, and thereby improves the efficiency of \hpo.

Our experiments with three \studies show that \stage-based execution can reduce GPU-hours and end-to-end training time compared to the \trial-based execution. When tuning learning rate, it can save end-to-end training time up to 2.98 times, and GPU-hours up to 5.73 times. When tuning batch size, it can save end-to-end training time up to 4.13 times and GPU-hours up to 6.60 times.

\section{Stage-based Execution}

\begin{figure}[tb]
    \centering
    \begin{subfigure}[b]{0.49\textwidth}
       	\centering
        \includegraphics[width=1.0\textwidth]{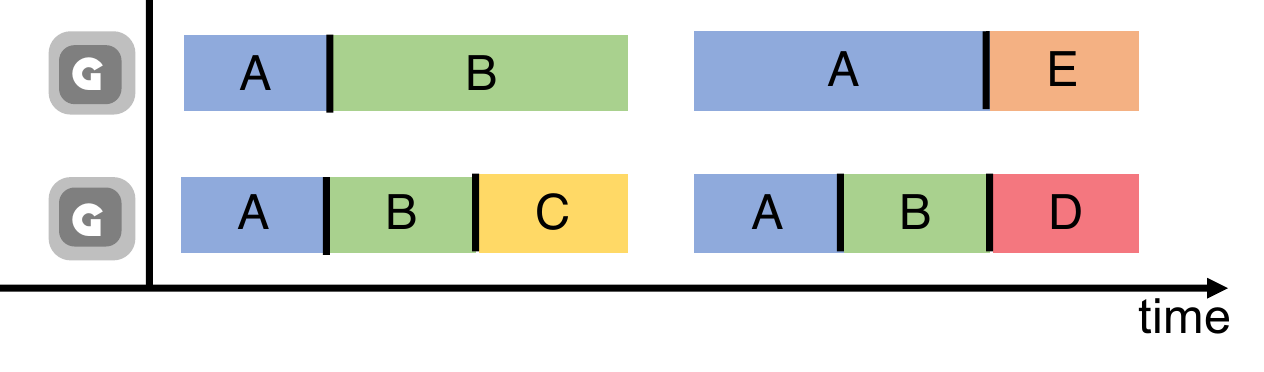}
        \centering
       	\caption{trial-based execution}
       	\label{fig:trial-exec}
    \end{subfigure}
    \hfill
    \begin{subfigure}[b]{0.49\textwidth}
        \includegraphics[width=1.0\textwidth]{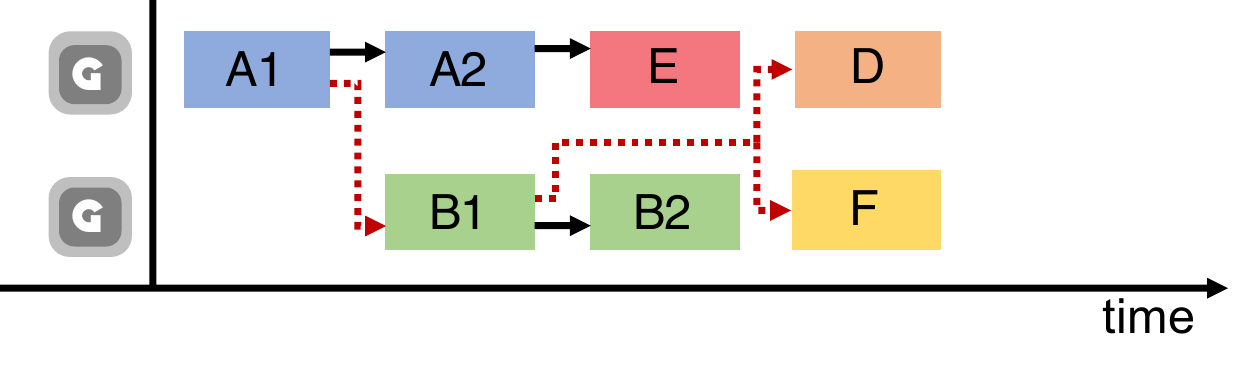}
    	\caption{stage-based execution}
    	\label{fig:stage-exec}
    \end{subfigure}
    \caption{
        We can execute the trials in Figure~\ref{fig:hp-sequence} in two ways: trial-based, and stage-based. Trial-based execution, shown on the left, executes each trial independently on a GPU. As there are two GPUs, each trial is assigned one GPU.
        Stage-based execution takes advantage of the stage-tree in Figure~\ref{fig:stage-tree}. The edges of the tree are shown in the figure as red dotted or black solid lines.
    }
    \label{fig:execution}
\end{figure}

We propose a \stage-based execution strategy for using GPU resources efficiently. In this section, we describe how a \stage-based system represents \trials internally, and how it executes \stages.

\subsection{Stage-tree}

Given a search space composed of multiple \trials as in Figure~\ref{fig:hp-sequence}, instead of directly executing each \trial, we express the search space as a \stage-tree, as in Figure~\ref{fig:stage-tree}.
In the example, there were four trials which were split and merged into seven stages.
In Figure~\ref{fig:stage-tree}, each node is a stage; it has homogeneous \hp values with a fixed range of iterations\footnote{Number of iterations can be number of epochs or steps depending on user code.}.
As there were four trials, there are corresponding four leaf stages in the stage-tree.
Therefore, every path from a root node to a leaf node corresponds to a \trial with a unique \hp sequence.

The \stage-tree is an internal representation, and therefore is not exposed to the user.
Users submit a study composed of \trials and they are automatically split into homogeneous configurations by the system.
Then, to add a trial into the stage-tree, the system traverses the stage-tree built so far to find the longest common prefix between the given \trial and the current stage-tree.
In case the number of training steps are different for two identical configurations, it splits the larger sequence to maximize the length of common prefix. Then, it appends the remaining unmatched subsequence of the new \trial to the tree.
Note that in this case each homogeneous value corresponds to a list of \stages, as in Figure~\ref{fig:stage-tree}.
The first homogeneous value (0.1) of Trial 1 and Trial 2 is identical, but as one sequence is longer than the other, it splits the longer one into \stages A1 and A2.

\subsection{Stage Execution}

To use GPU resources more efficiently and reduce end-to-end time, we employ stage-based execution based on the stage-tree built from trials.
With trial-based execution, we treat each trial as a black box and simply launch the trial on an idle GPU, as shown in Figure~\ref{fig:trial-exec}.
On the other hand, when we use the stage-based execution strategy, each \stage becomes a unit of scheduling, thus avoiding redundant computation.

Figure~\ref{fig:stage-exec} shows an example of stage-based execution.
We can see that stage-based execution improves both GPU-hours and end-to-end time compared to trial-based execution.
In the figure the edges indicate a parent-child relationship between the two connected nodes in the stage-tree.
Each child node starts execution by initializing its weights using the checkpoint of its parent node.
Therefore, each edge represents a data dependency relationship; every child starts off from where its parent has ended.
For the red dotted lines, initializing weights from a checkpoint is necessary. However for the black solid lines, since the connected nodes run consecutively in the same GPU, checkpoint loading is unnecessary.
For example, executing A1 and A2 consecutively does not require checkpoint loading, but executing E and D in series in the same GPU needs loading of checkpoint from B1 before running D.

We keep track of \stages in the tree that have been executed and are currently being executed.
The children of \stages that have been executed become candidates for execution.
When deciding the next \stage to execute among the candidates, we consider the priorities of \trials that many \hp optimization algorithms specify. For example, existing algorithms such as the Asynchronous Successive Halving Algorithm (ASHA)\cite{massively} not only specify what configuration to run, but also in what order.
Given the next \stage to execute, we decide how many GPUs the \stage needs to be executed.
In order to do so, we profile the GPU usage characteristics of \stages and estimate the resource requirements of new \stages based on the history of previously executed \stages.
If the stage fails because our prediction turns out to be wrong, we then re-launch the stage by increasing
the number of GPUs until we find a sufficient number of GPUs to execute the \stage.

Before training each \stage, the system spawns multiple workers, each responsible of one or more GPUs, and run inside a containerized environment.
Each container exclusively holds multiple GPUs from the same node, and runs one worker process. 
Note that \trials from a single \study share a common environment, hence it is perfectly valid to reuse both the container and the worker process.
As a result, we can make the system not launch new containers or workers for \stages with the same resource requirements. When running a \stage with different resource requirements, we need to merge or split existing workers.
For example, if all existing workers only have one GPU in control, and the next \stage requires two GPUs, the system destroys two free workers (and containers) and creates a new worker with two GPUs.
To avoid communication overhead in distributed training, the system tries to select and merge workers from the same node. 

\section{Discussion}

\subsection{Just-fit Resource Allocation}
By splitting \trials into \stages, besides the advantage of reducing computation, the system is able to provide high resource utilization by allocating the right amount of resource. DL training jobs require varying amounts of resources according to the \hps, and \trials require different amounts of resources through its life cycle with respect to its \hp sequence.

For instance, when training a model using dynamic batch size~\cite{dont-decay-lr} or optimization algorithms~\cite{adam2sgd}, the memory requirements of a \trial vary. In such situations, \trial-based execution should allocate the maximum amount of resources the \trial may use, which incurs lots of idle resources. On the contrary, we can avoid such inefficiency by allocating GPU resources for each stage.
Since each \stage has a homogeneous behavior, we are able to allocate just the right amount of resources to each \stage.

\subsection{Multi-study Optimization}

Multiple \studies with an identical model and dataset can run on the same DL cluster \cite{multi_tenant}. 
In this case, if we use \stage-based systems, we can expose the search space history explored by previous study to other \studies. Then, \trials from one \study may exploit the search space that other \studies have already explored by reusing the results. {\Hpoa}s that require prior information to work properly can benefit from using the search space of previous runs.
\revise{To support multi-study optimization in our system, we can use an executor layer. There are a set of studies and a set of executors, and the system maps each study to an executor. Studies that have the same dataset and model are routed to the same executor. A router component will send submitted trials to its corresponding executor based on which study it came from. This way, we can give each executor its stage-tree and resource pool, as merging will not occur between executors.}

\subsection{Supporting Continuous Search Space}

For our stage-based execution system to perform well, the \trials sampled by \hpoas should have overlapping prefixes. If the user declares \hp search space as a set of discrete values (e.g., grid search), we have overlapping prefixes more or less depending on the sampling algorithm.
However, when using random search over a continuous domain, sampled values are hardly ever identical.
In other words, the possibility of benefiting from overlapping prefixes is close to zero.
This is a typical case in random search or bayesian optimization over a continuous domain, where our system may provide only minimal gain.

In cases when there are no overlaps among trials, our system has identical behavior with previous systems. For example, when running algorithms like random search or bayesian optimization, sampled trials can have very low overlaps, and such algorithms do not hinder system performance. Rather, our system gives optimization algorithms and users an opportunity to reduce computation in training many trials.
The user can further modify the search space in a way that maximizes prefix overlaps, as more overlaps will allow the system to explore the space better within the same time or financial budget.
One may think these constraints improves efficiency at the cost of the final accuracy.
However, improving the time- or resource- efficiency of \hpo is also important for improving the quality of the final model.
Since we can try training with more \hp configurations within the same budget, the final model may actually become better than before, and being able to continue training from existing checkpoints will help in finding network weights that perform high validation accuracy. 
Indeed, algorithms like random search or bayesian optimization are designed with the same philosophy in mind; they selectively choose \hps to explore the search space efficiently.

\section{Experimental Evaluation}
\label{evalaution}

To evaluate our stage-based execution strategy, we implemented a prototype system in Python. We use Docker~\cite{docker} to launch containers and gRPC~\cite{grpc} as a messaging interface between processes. Estimation of GPU requirements for each stage is done by a simple linear regression model, yet the estimation falls back to one GPU at cold start.

We present three experiments that show how \stage-based execution benefits in \hpo.
We apply both \trial-based and \stage-based execution strategies to optimize \hps and measure the required GPU-hours and end-to-end time.
End-to-end time refers to the elapsed time from the experiment's start to end, while GPU-hours signifies the sum of active execution time of all GPUs (e.g., executing a job on 2 GPUs for 10 hours results in 20 GPU-hours).
The first two experiments involve tuning the learning rate, and the last experiment involves tuning the batch size. All three cases have reached target accuracy.

The first two cases is training the ResNet-20~\cite{resnet} model on the CIFAR-10 dataset~\cite{cifar}. Since the CIFAR-10 dataset does not have a fixed \textit{train-validation-test} split, we used a 40K/10K/10K split; 40K images are used to train the model, 10K images are used to tune the \hps, and the remaining 10K images are used to report the final test error on the model with the lowest validation error. For each case, we used grid search and the Successive Halving Algorithm (SHA) \cite{jamieson2016non} to optimize the \hps. SHA cuts down running trials by a factor of three at epoch 16 and 64 with their validation accuracy.

As the dimension of \hp sequences is too large to search, optimizing the \hp sequence directly is not practiced. Instead, it is common to parameterize the sequence and set the parameters as {\hp}s. For example, the learning rate is often modeled as a step-decay function and the epoch to decay the learning rate is optimized instead~\cite{NAS-RL, massively}.
In addition, to take full advantage of computation reuse, we approximated the continuous \searchspace of \hps into a discrete \searchspace.
The \hps used are shown in Table~1(a). We only tune the learning rate, and there are three decay periods. Each trial decreases the learning rate three times. Each decay period represents the number of epochs until the next decay.
The \trial-based and \stage-based setups both explored 108 \trials that originated from the same \hp \searchspace. All trials are given maximum 200 epochs of training, unless early stopped.

\begin{table}[tb]
  \centering
  \begin{subtable}[t]{0.45\textwidth}
    \centering
    \begin{tabular}{c | c}
        \toprule
        \Hp & Values \\
        \toprule
        initial learning rate & 0.5, 0.2 \\
        decay rate & 0.2, 0.1 \\
        decay epoch 1 & 40, 60, 80 \\
        decay epoch 2 & 40, 60, 80 \\
        decay epoch 3 & 40, 60, 80 \\
        batch size & 128 \\
        optimizer & SGD \\
        momentum & 0.9 \\
        weight decay & 1e-4 \\
        \bottomrule
    \end{tabular}
    \caption{Resnet}
    \end{subtable}
        \hfill
    \begin{subtable}[t]{0.45\textwidth}
        \centering
        \begin{tabular}{c | c}
            \toprule
            \Hp & Values \\
            \toprule
            initial batch size & 128 \\
            increase rate & 5 \\
            increase epoch 1 & 30, 60 \\
            increase epoch 2 & 30, 60 \\
            increase epoch 3 & 30, 60 \\
            learning rate & 0.1 \\
            optimizer & SGD \\
            momentum & 0.9 \\
            weight decay & 5e-4 \\
            \bottomrule
        \end{tabular}
        \caption{WideResnet}
    \end{subtable}
  \caption{\Hp space}
  \label{tbl:hpconf}
  \centering
\end{table}

The experiments were conducted on 4 machines with total 20 NVIDIA GeForce TITAN Xp GPU cards and 2 18-core Intel Xeon E5-2695 @ 2.10 GHz processors with 256 GB RAM.
For both \trial-based and \stage-based experiments, we measure the end-to-end time and GPU-hours (total GPU resource usage) as shown in Figure~3.

The reported test error of ResNet-20 on CIFAR-10 is 8.75\% \cite{resnet}. Our best model's test error reaches 8.24\% with only 4/5 of original training data. As shown in Figure~3, for grid search, stage-based system is 2.94 times faster and uses 3.49 times less resource.
The SHA has a diminishing effect in reducing end-to-end time. This is because most of the GPUs were idle in stage-based execution. In fact, the GPU resource usage is 5.73 times smaller than \trial-based execution.

\begin{figure}[tb]
    \centering
    \begin{subfigure}[b]{0.49\textwidth}
       	\centering
        \includegraphics[width=1.0\textwidth]{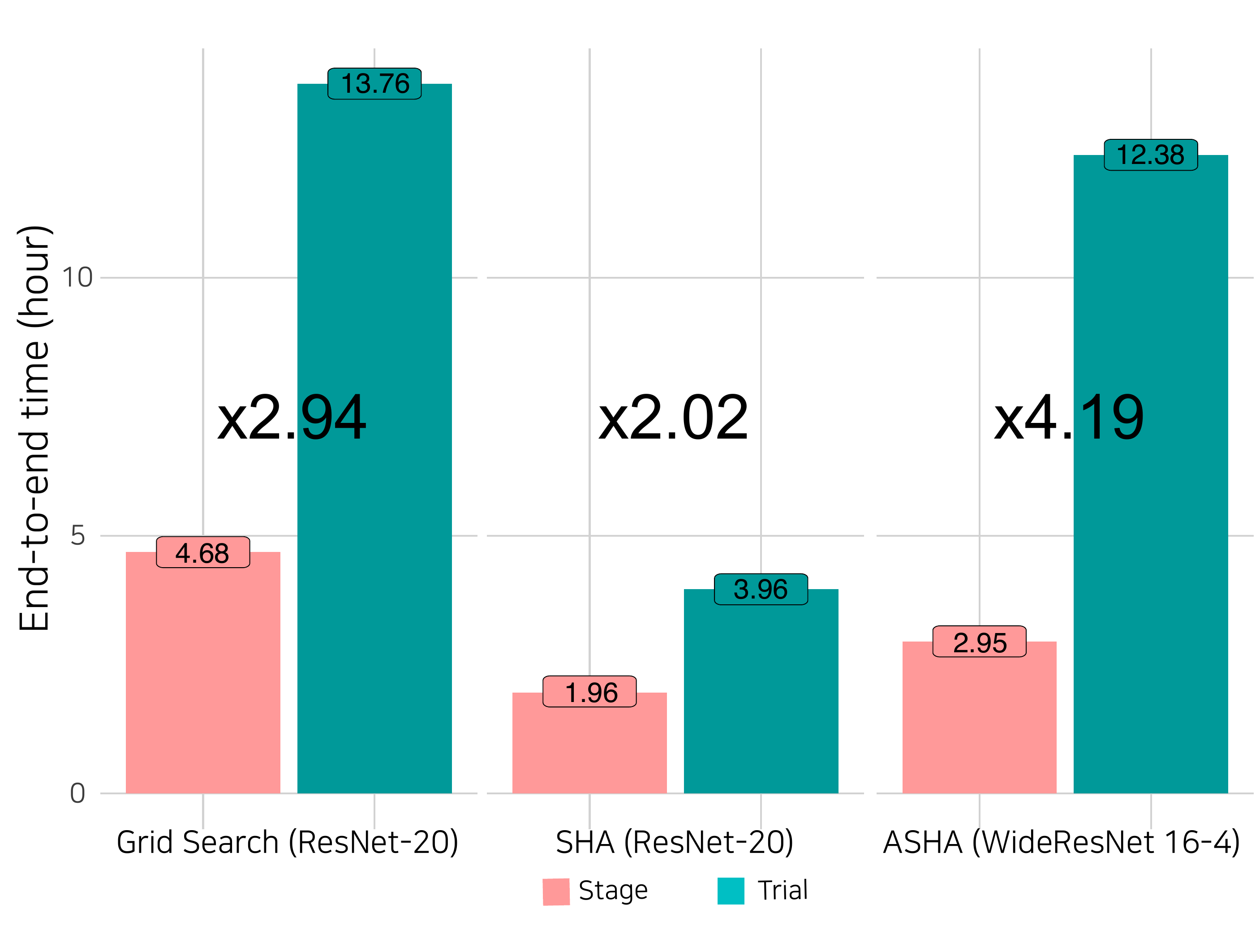}
        \centering
       	\caption{End-to-end time}
       	\label{fig:e2e}
    \end{subfigure}
    \hfill
    \begin{subfigure}[b]{0.49\textwidth}
        \includegraphics[width=1.0\textwidth]{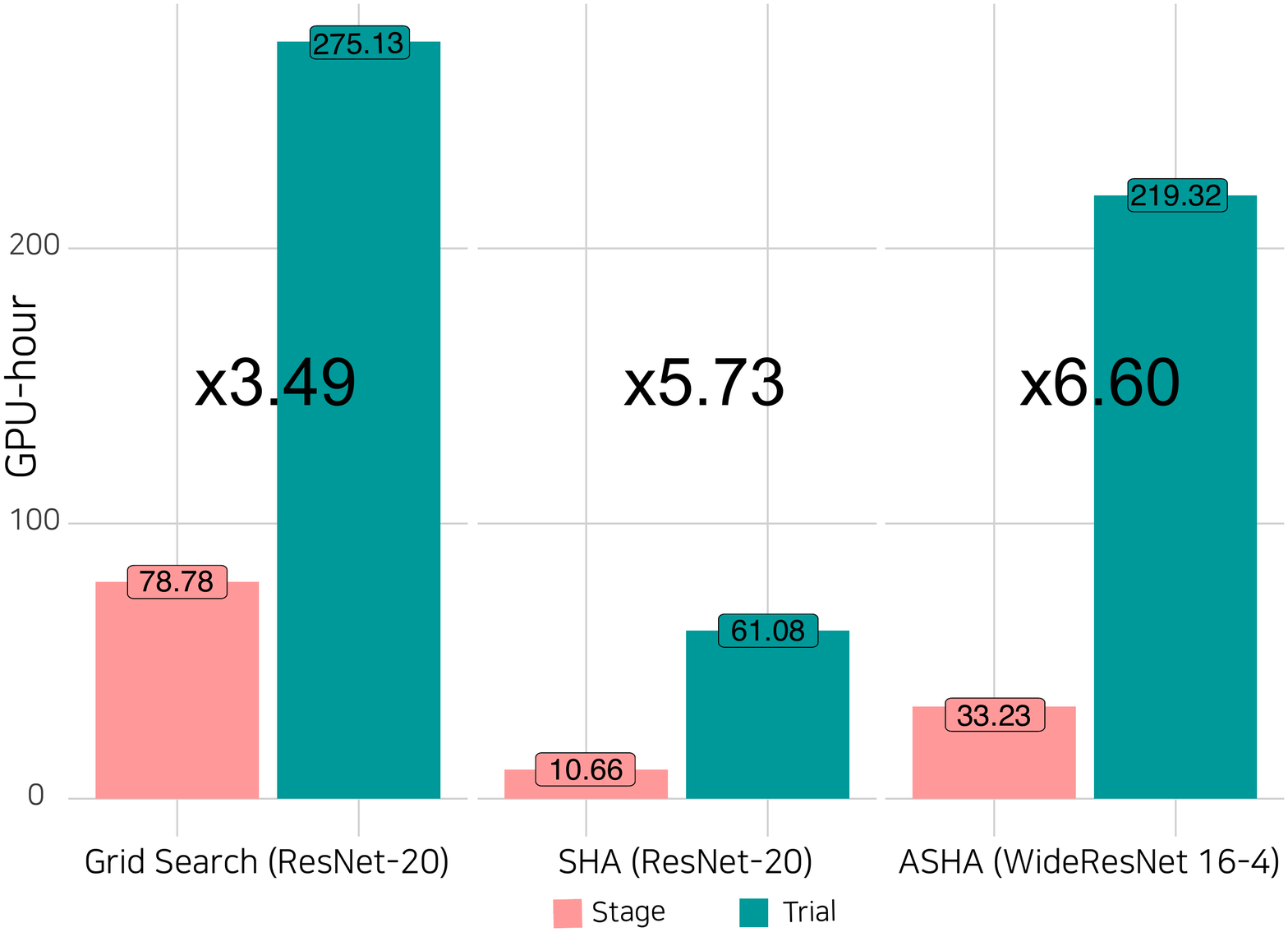}
    	\caption{GPU-hour}
    	\label{fig:gpu}
    \end{subfigure}
    \caption{
        Experiment results
    }
    \label{fig:eval}
\end{figure}

The third case involves just-fit resource allocation, which was discussed in Section~3.1. We have trained the \textit{WideResnet 16-4} model on the CIFAR-10 dataset. We only tuned batch size; initial value, which epoch to increase, and by how much as in Table~1(b). Total 64 trials were run for both \trial-based and \stage-based settings. We use the same \textit{train-validation-test} split on CIFAR-10.
Grid search and the ASHA\cite{massively} algorithm was used to tune the model. Parameters $R=216, r=15, eta=3, s=0$ are used to run the ASHA algorithm.
The results show the validation accuracy reaches 94.8\%, where the paper that introduces this strategy\cite{dont-decay-lr} has reached 94.4\%. In addition, by assigning resources per stage, stage-based approach reduces resource spendings by 6.6 times.

\section{Conclusion}

In this paper, we have proposed the \stage-based execution strategy that splits \trials into smaller homogeneous units and removes computational redundancy in the \hpo process.
Applying this strategy to \hpo, we are able to reduce end-to-end training time and GPU-hours by up to 4.19 times and 6.6 times, respectively.

As a future work, we plan to evaluate this execution strategy in various state-of-the-art models and datasets using various \hps. \revise{In addition, in this work, we evaluated only discrete-valued sequences. We will expand our research to continuous-valued sequences as well as \hps such as data augmentation or network architecture parameters.}
Furthermore, we plan to develop a new \hpoa that can maximize the use of this strategy.

\subsubsection*{Acknowledgments}

This work was supported by Institute for Information \& communications Technology Promotion (IITP) grant funded by the Korea government (MSIT) 
(No.2015-0-00221, Development of a Unified High-Performance Stack for Diverse Big Data Analytics),
the ICT R\&D program of MSIT/IITP (No.2017-0-01772, Development of QA systems for Video Story Understanding to pass the Video Turing Test),
and Samsung Advanced Institute of Technology.

\bibliographystyle{plain}
\bibliography{hippo}
\end{document}